\pdfoutput=1
\documentclass[11pt]{article}

\usepackage{acl}

\usepackage{times}
\usepackage{latexsym}
\usepackage{subcaption}
\usepackage{amsmath}
\usepackage{arabtex}
\usepackage{utf8}
\setcode{utf8}
\usepackage{CJKutf8}

\usepackage{multirow}

\usepackage{microtype}
\usepackage{graphicx}
\newcommand{\datasetname}{ArtELingo}

\begin{document}

\title{\datasetname{}: A Million  Emotion Annotations of WikiArt  with \\
 Emphasis on Diversity over Language and Culture}

\author{{\bf Youssef Mohamed}\textsuperscript{1}\thanks{\quad Corresponding Authors}\quad {\bf Mohamed Abdelfattah }\textsuperscript{1}\quad {\bf Shyma Alhuwaider}\textsuperscript{1} \quad {\bf Feifan Li }\textsuperscript{1} 
\vspace{1mm} \\
{\bf Xiangliang Zhang }\textsuperscript{2} \quad {\bf Kenneth Ward Church }\textsuperscript{3} \quad {\bf Mohamed Elhoseiny}\textsuperscript{1}\textsuperscript{*} 
\vspace{1mm} \\
\textsuperscript{1}KAUST \quad \textsuperscript{2} University of Notre Dame \quad\textsuperscript{3} Northeastern University
\vspace{1mm} \\
\fontsize{9}{9} \selectfont \texttt{\{youssef.mohamed,mohamed.abdelfattah,shyma.alhuwaider,feifan.li\}@kaust.edu.sa}\\ \fontsize{9}{9} \selectfont \texttt{ xzhang33@nd.edu, k.church@northeastern.edu, mohamed.elhoseiny@kaust.edu.sa}}

%%%%%%%%%%%%%%%
% TEASER START
%%%%%%%%%%%%%%%
\makeatletter
\let\@oldmaketitle\@maketitle
\renewcommand{\@maketitle}{\@oldmaketitle
\myfigure\bigskip}
\makeatother
\newcommand\myfigure{%
\vspace{-4mm}
%   \makebox[0pt]{\hspace{0cm}\includegraphics[width=\linewidth]{figures/figure_no_ws (6).pdf}}
  \includegraphics[width=\linewidth]{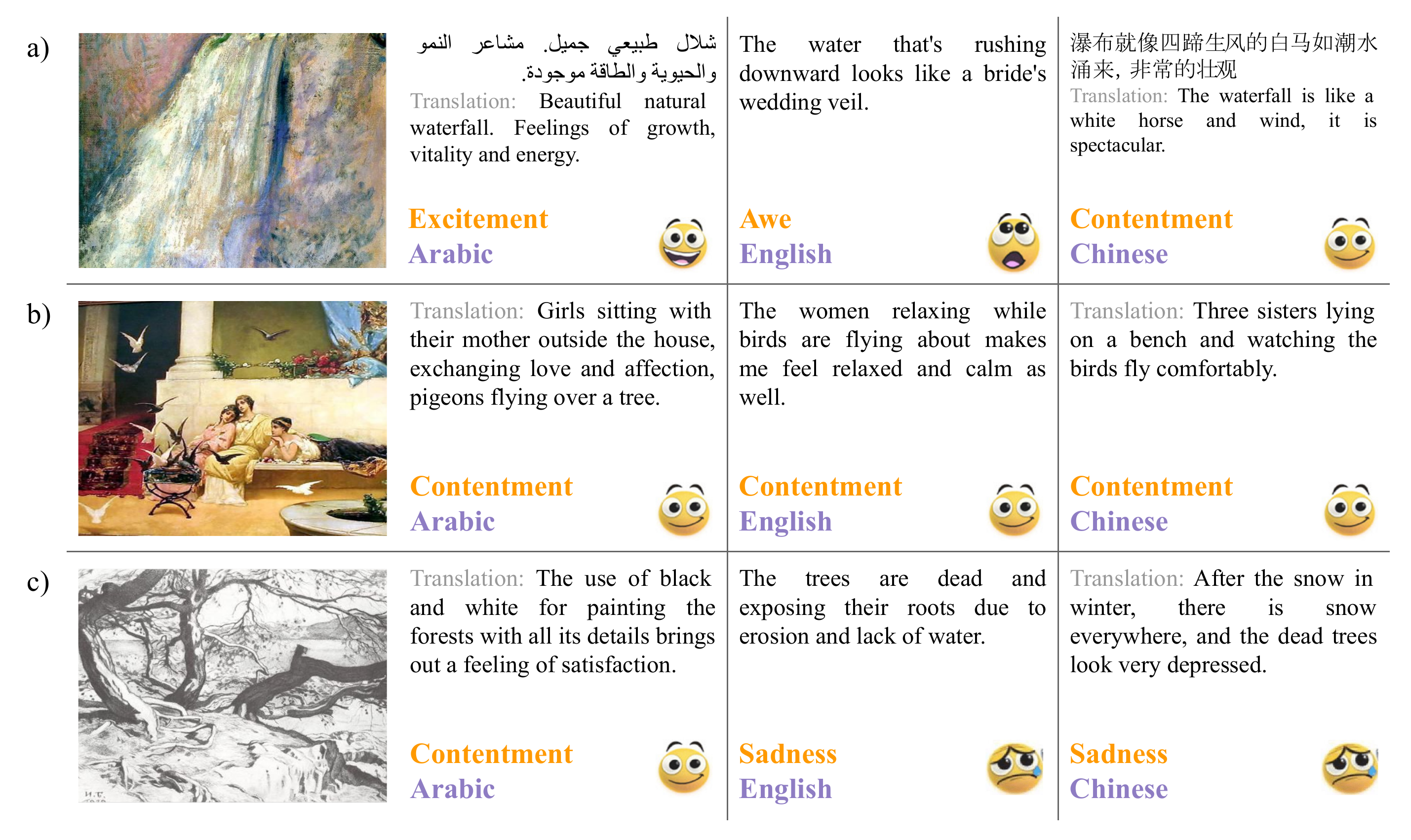}
 \\
  \centering
  \refstepcounter{figure}\normalfont{Figure~\thefigure: \datasetname{}, a multilingual dataset  and benchmark of
  WikiArt with captions \& emotions
  } 
  \label{fig:teaser}
}

% \makeatother

%%%%%%%%%%%%
% TEASER END
%%%%%%%%%%%%
\maketitle

\begin{abstract}

This paper introduces \datasetname{}, a new benchmark and dataset,
designed to encourage work on diversity across languages and cultures.
Following ArtEmis, a collection of 80k artworks from WikiArt with 0.45M emotion labels and English-only captions,
\datasetname{} adds another 0.79M annotations in Arabic and Chinese, plus 4.8K in Spanish to evaluate ``cultural-transfer'' performance.
More than 51K artworks have 5 annotations or more in 3 languages.
This diversity makes it possible to study similarities and differences across languages and cultures.  Further, we investigate captioning
tasks, and find diversity improves the performance of baseline models.
\datasetname{} is publicly available\footnote{\url{www.artelingo.org}} with standard splits and baseline models.
We hope our work will help ease future research on 
multilinguality and culturally-aware AI. 

\end{list}
\end{abstract}

\section{Introduction}
\label{sec:intro}

Figure~\ref{fig:teaser} compares and contrasts annotations on WikiArt across language/culture.
We believe these differences are interesting and important, and far from random.
One might suggest using machine translation to translate English captions to many other languages,
but we believe that doing so would miss much of the opportunity. 
Building human-compatible AI that is more aware of our emotional being is important for increasing the social acceptance of AI. ArtEmis \cite{achlioptas2021artemis} is an important step in this direction, introducing a collection of 0.45M emotion labels and affective language explanations in \emph{English} on more than 80,000 artworks from WikiArt. However, by design, ArtEmis is limited to
English, lacking coverage of other cultures and languages.

Cultural differences are a major source of diversity~\cite{meyer_2014}. The customs, social values, lifestyles, and history of different countries and cultures greatly influence human behavior.  Emotional experiences are no exception; people from different countries respond differently to similar scenarios. For example, 
a person born and raised in a Nordic country would be more comfortable in a lush forest than in a desert,
but a Bedouin may be more comfortable in a desert than in a forest.

Consider Figure~\ref{fig:teaser}c,
where an Arabic annotator assigned the image the label \textbf{contentment},
but the other two annotators used the label: \textbf{sadness}.
Captions are useful for diving deeper into these differences.
The \textbf{sadness} annotations mention \textit{death}\footnote{ \begin{CJK*}{UTF8}{gbsn}冬天下雪后到处白雪皑皑，枯树显得很萧条。 \end{CJK*} (snow everywhere and dying trees is depressing)} and disasters,\footnote{\textit{no me gusta el ambiente, lo primero que me vino a la mente fué un desastre natural con destrucción a su paso} (mentions a natural disaster)}
in contrast with the  \textbf{contentment} annotation that ends
with: \textit{feeling of satisfaction}.

There can be interesting differences between languages/cultures even
when annotators use the same label.
Consider Figure~\ref{fig:teaser}b, where
all three labels are \textbf{contentment}.  Although the three captions agree
on the label, two of the captions imply that some/all of the girls are sisters,
but there is no such implication in the English caption.

We believe deep nets will be viewed as more culturally aware,
if they can 
capture linguistic/cultural patterns such as these.
Emotions are based on past experience, and play an integral role in determining human behavior. Not only they reflect our internal state but also directly effect how we perceive, interpret external stimuli~\cite{izard2009emotion}, and how to act based on them~\cite{lerner2015emotion}. Hence, studying emotions is essential to exploring a confounding aspect of human intelligence. 

In summary, our \textbf{contributions} are:
\begin{enumerate}
  \setlength{\itemsep}{0pt}
  \setlength{\parskip}{0pt}
 \setlength{\parsep}{0pt}
    \item 0.79M annotations (labels + captions) in Arabic and Chinese, plus 4.8k in Spanish,
    \item a benchmark with standard splits, and
    \item baseline models for two tasks: (1) label prediction and (2) affective caption generation.
\end{enumerate}

The rest of the paper is organized as follows: related work is discussed in \S\ref{sec:related},
followed by our main motivation in \S\ref{sec:motivation}, and data collection in \S\ref{sec:artemis}.  \S{\ref{sec:analysis}} provides qualitative and quantitative analyses of \datasetname{}.  Baseline models for emotion label prediction and caption generation are presented in \S\ref{sec:emo-cls} and \S\ref{sec:neural-speakers}, respectively.

\begin{table}[hb!]
\centering
\begin{tabular}{ r | c c c}
  \setlength{\tabcolsep}{1pt}
 & {COCO} & {ArtEmis} & {\datasetname{} }\\ \hline
\small{{Image Source}} & Photos & WikiArt & WikiArt \\
\small{{\#Images}} & 328k & 80k & 80k \\
\small{{\#Annotations}} & 2.5M & 0.45M & 1.2M \\
\small{{\#Annot/Image}} & 7.6 & 5.68 & 15.3 \\
\small{{Emotions}} & 0 & 9 & 9 \\
\small{{Languages}} & E & E & ACES \\ \hline
\end{tabular}
\caption{\label{tab:benchmarks} A Comparison of Three Datasets.  \datasetname{} has a million annotations from ACES: Arabic (A), Chinese (C), English (E) and Spanish (S).}
\end{table}

\begin{figure}
  \centering
  \begin{subfigure}{0.46\linewidth}
    \includegraphics[width=\linewidth]{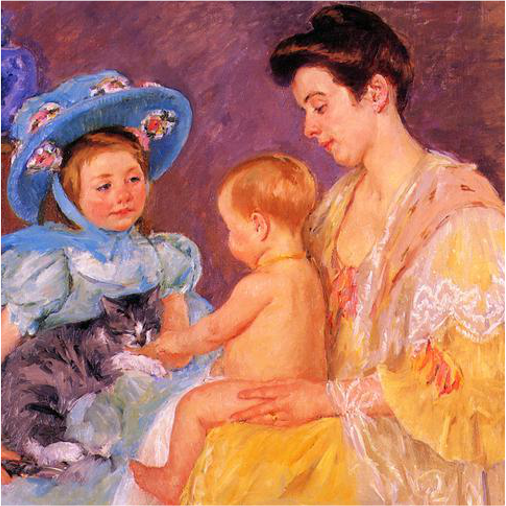}
    \caption{ArtEmis: {
    \textit{I love everything about this painting of a mother and her two children lovingly interacting with the family pet cat.}}}
    \label{fig:familya}
  \end{subfigure} \hspace{2mm}
  \begin{subfigure}{0.46\linewidth}
    \includegraphics[width=\linewidth]{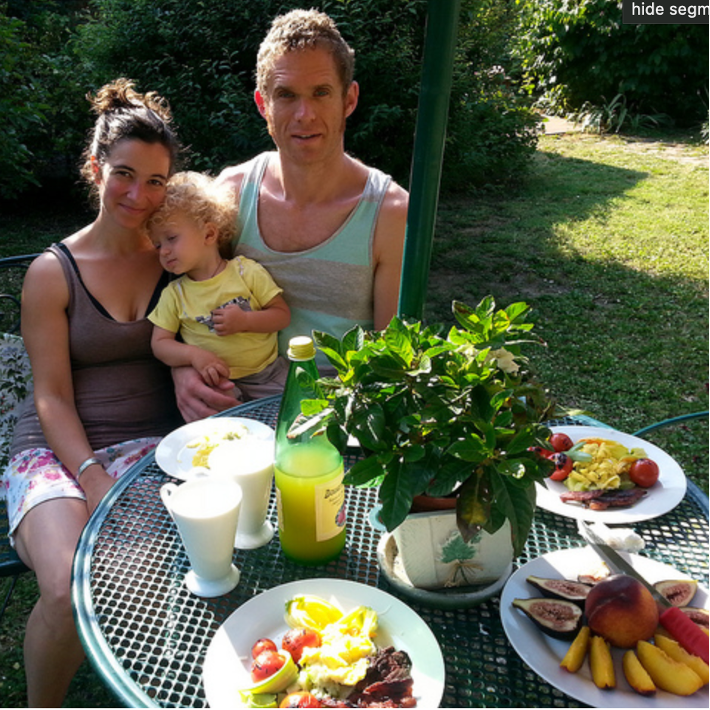}
    \caption{COCO: \textit{A man and a woman holding a little kid while sitting at a table outside\\}}
    \label{fig:familyb}
  \end{subfigure}
  \caption{\label{fig:family} COCO captures the facts, and ArtEmis enhances those facts with emotion/commentary.}

\end{figure}

\section{Related Work}
\label{sec:related}

\subsection{Captions with Emotions}
\label{sub_sec:art_cap_emo}

Work on captioning is moving beyond factual captions in early benchmarks such as
COCO \cite{lin2014microsoft}.
Figure~\ref{fig:family} shows two images of families, one from 
ArtEmis and the other from COCO.
Both captions capture the facts, but ArtEmis enhances the facts with emotion/commentary.

\begin{figure}[ht]
    \includegraphics[width=\linewidth]{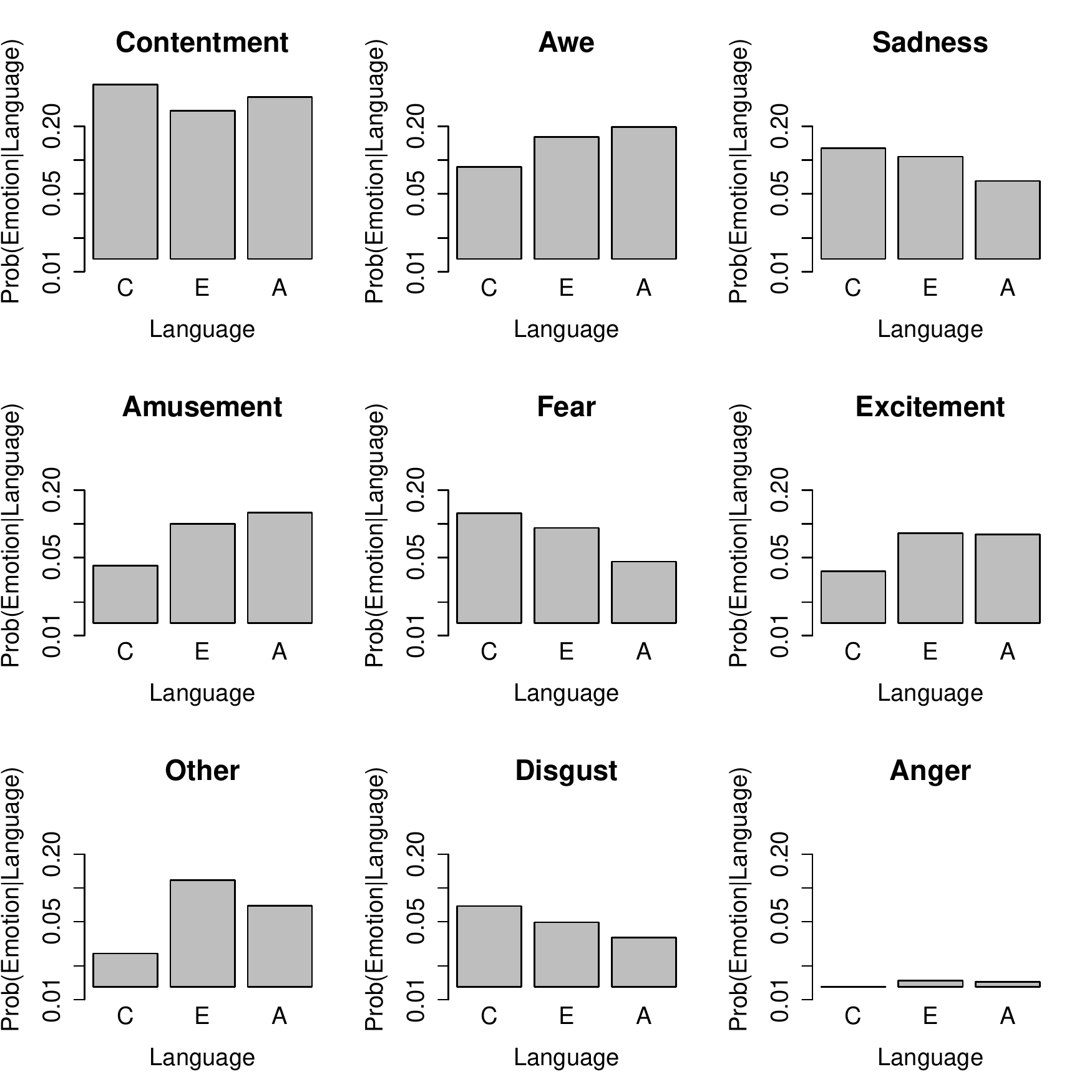}
    \caption{In general, positive emotions are more frequent than negative emotions.
    The 9 plots are sorted by probability.  The log scale on y-axis highlights relative probabilities.
    A (Arabic) is relatively high for some classes (\textbf{awe}), and low for others (\textbf{sadness}).}
    \label{fig:box9}
\end{figure}

Table~\ref{tab:benchmarks} compares three benchmarks: COCO \cite{lin2014microsoft}, ArtEmis and \datasetname{}.
Art\underline{E}mis encourages work on emotions by replacing COCO photos with WikiArt,\footnote{\url{https://www.wikiart.org/}}
and by introducing 9 emotion classes, 4 positive,\footnote{Positive: \textbf{Contentment}, {\textbf{Awe}}, {\textbf{Amusement}}, {\textbf{Excitement}}}
4 negative\footnote{Negative:
{\textbf{Sadness}}, {\textbf{Fear}}, {\textbf{Disgust}}, {\textbf{Anger}}} and {{Other}}.
\datasetname{} encourages researchers to work on visually grounded multilinguality by providing affective annotations
in three languages (henceforth, ACE/ACES): Arabic, Chinese and English. In addition, we provide a small set of Spanish (S).
Figure~\ref{fig:box9} shows that positive emotions are more frequent than negative emotions,
especially in Arabic.

\subsection{Related Work in Other Fields}

There is a considerable literature on emotions, especially in Psychology \cite{russell1999core}.
One can find quite a few benchmarks on emotion in HuggingFace:
\cite{saravia-etal-2018-carer,demszky-etal-2020-goemotions,xiao2018mes}.\footnote{\label{hug_face}\url{https://huggingface.co/datasets?sort=downloads&search=emotion}}
There are a number of papers in computational linguistics on emotion and Chinese \cite{chen-etal-2020-end,quan-ren-2009-construction,wang-etal-2016-bilingual,lee-etal-2010-emotion},
and on emotion and Arabic \cite{abdullah-shaikh-2018-teamuncc}.
There is also considerable work on emotion in other fields such as vision \cite{Mittal_2021_CVPR}.

Many datasets have been collected to study
emotional responses to modalities such as:

\begin{itemize}
  \setlength{\itemsep}{0pt}
  \setlength{\parskip}{0pt}
 \setlength{\parsep}{0pt}
    \item Text
    \cite{strapparava-mihalcea-2007-semeval,demszky-etal-2020-goemotions,mohammad-etal-2018-semeval,liu2019dens},\footnote{\url{https://data.world/crowdflower/sentiment-analysis-in-text}} 
    
    \item Image~\cite{mohammad2018wikiart,Kosti_2017_CVPR_Workshops}, and 
    \item Audio~\cite{cowen2019mapping,cowen2020music}. 
\end{itemize}

Bias is the flip side of inclusiveness.
There
has been considerable discussion recently about biases \cite{bender2021dangers,NIPS2016_a486cd07,buolamwini2018gender,10.1145/3457607, liu2021visually}.
Some of this work is more relevant to our interest in Chinese \cite{jiao-luo-2021-gender,liang-etal-2020-monolingual},
and Arabic \cite{abid2021persistent}.
Many machine learning methods will, at best, learn what is in the training data.
There have been some attempts to remove biases in corpora,
but it might also be constructive to create more inclusive benchmarks such as \datasetname{}.

Awareness of different cultures is becoming increasingly important. Gone are the days when it was sufficient for datasets to focus on a single culture. Recently, the Vision \& Language community has been producing more multicultural multilingual datasets \cite{bugliarello2022iglue, srinivasan2021wit, armitage2020mlm}. \datasetname{} contributes cultural diversity over emotional experiences. The effect of culture on psychology has been studied in separate studies \cite{henrich2010weirdest, abu1990romance, norenzayan2005psychological}. \datasetname{} provides empirical evidence that might motivate cultural psychology studies.

\section{Opportunities for Improvement}
\label{sec:motivation}

Many of the resources mentioned above have advanced our understanding of the relationship between emotion and various stimuli, through there are
always opportunities for improvement.  We are particularly interested in three such opportunities:
scale, multimodality and multilinguality/muli-culturalism.  
As for scale, demand for  larger training sets is expected to continue to increase, 
given  the rise of large scale foundation models~\cite{bommasani2021opportunities}.

As for multimodality, although most benchmarks mentioned above focus on a single modality,
there are a few multimodal exceptions such as IEMOCAP \cite{busso2008iemocap}, COCO and ArtEmis.
IEMOCAP collected speech and facial and hand movements of 10 actors.  Unfortunately, this approach may be  expensive to scale up.

The use of Amazon Mechanical Turk in ArtEmis is easier for scaling, however,
ArtEmis is limited to English.   \datasetname{} addresses
multilinguality/multi-culturalism by adding Arabic and Chinese annotations.
We use languages as a proxy to reflect different cultures.
English is a representative sample of the West, and Chinese is a representative sample of the East,
and Arabic is a representative sample of the Middle East.

\subsection{Representation of Regions in WikiArt}
\label{sec:representation}

\begin{table}
    \centering{
        \begin{tabular}{l | r  r}
            Region     &   \#Artworks & \%  \\ 
            \hline  
            West (Non English) &142.8k&57.1\%\\
            West (English) & 54.0k & 21.6\% \\
            Other & 38.0k & 15.2\% \\ 
            Middle East (Non Arabic) & 12.2k & 4.8\% \\
            Middle East (Arabic) & 1.6k & 0.6\% \\
            
            East (Chinese) &1.4k & 0.5\% \\
            \hline \hline
            Total & 250.0k & 100\% \\ 
            
        \end{tabular}

    \caption{WikiArt is more representative of the West}
    \label{tab:wikiArt_diversity}}
\end{table}

\datasetname{} assumes that WikiArt is a representative sample of the cultures of interest.
While WikiArt is remarkably comprehensive, Table~\ref{tab:wikiArt_diversity} suggests
the WikiArt collection has better coverage of the West than other regions of the world.
This table is based on WikiArt's assignment of artworks to nationalities.\footnote{\url{https://www.wikiart.org/en/artists-by-nation}}
We assigned each nationality to West (English\footnote{West (English):\textit{ Americans, Australians, British} and \textit{Canadians}} 
and Non English\footnote{West (Non English): \textit{Albanians, Armenians, Austrians, Azerbaijanis, Belarusians, Belgians, Bosnians, Bulgarians, Croatians, Czechs, Dutch, Estonians, Finnish, French, Georgians, Germans, Greeks, Hungarians, Icelandic, Irishes, Indigenous North Americans, Italians, Kazahstani, Latvians, Lithuanians, Luxembourgers, Maltese, Montenegrins, Polish, Portuguese, Romanians, Scottish, Slovaks, Serbians, Slovenians, Spanish, Swiss, Swedish, Ukrainians, Uruguayans, Uzbek} and \textit{Venezuelans}}), Middle East (Arabic\footnote{Middle East (Arabic): \textit{Algerians, Bahraini, Egyptians, Emiratis, Moroccans, Libyans, Lebanese, Iraqi, Palestinians, Qatari, Saudis, Syrians} and \textit{Tunisians}}
and Non Arabic\footnote{Middle East (Non Arabic): \textit{Kenyans, Jewish, Israeli, Iranians} and \textit{Turkish}}), East (Chinese) and Other.

\section{\datasetname{}}
\label{sec:artemis}

Following ArtEmis, we employ Amazon Mechanical Turk (AMT) platform to collect our data using interfaces ( see Figures ~\ref{fig:ar_interface},~\ref{fig:ch_interface},~\ref{fig:sp_interface} in the appendix). We faced a lack of Arabic and Chinese speaking annotators on AMT which led us to devise different strategies to recruit annotators. Arabic speakers were recruited by advertising the task in middle eastern universities encouraging students and their families to join our data collection efforts. Whereas Chinese speakers were recruited through Baidu who we'd like to thank. 

Annotators are asked to carefully examine each artwork before selecting the dominant emotion induced by it from a list of four positive, four negative emotions, and \emph{Other} to indicate a different emotion. Annotators are then asked to write captions that reflects the content of the artwork and explains their choice of emotion. Similar to ArtEmis, we collect annotations from five annotators for each artwork. 

\begin{figure}
    \centering
    \includegraphics[width=0.8\linewidth]{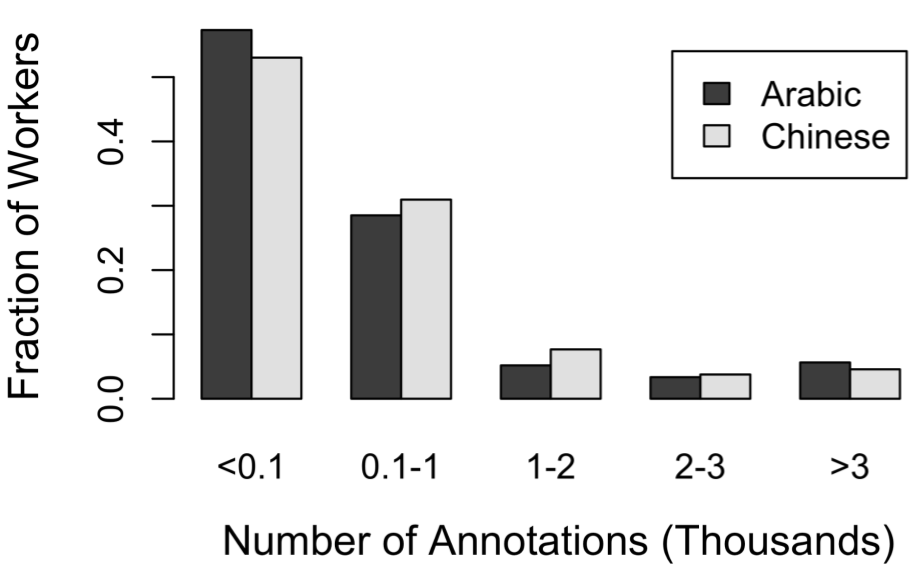}
    \caption{Most ($>$60\%) annotations are from long tail (workers who annotated less than 1K artworks).}
    \label{fig:workers}
\end{figure}

For a better cultural representation in \datasetname{}, we restrict the collection of different languages annotations to countries with large numbers
of native speakers. Chinese data is collected from China. For Arabic, we collect our data mainly from Saudi Arabia and Egypt. Finally, Spanish is collected from Latin America and Spain.
Figure~\ref{fig:workers} shows that most of the annotations are from a long tail of workers who
annotated less than 1000 artworks ensuring a diverse representation of cultures.

\noindent \textbf{Quality Control.} Annotations were rejected if they are too short, or if they are too similar to captions for other artworks.
In addition, 
a manual review was conducted by multiple reviewers, ensuring captions reflect the selected emotion label and the details of the artwork.
Table~\ref{tab:stats} reports some statistics on annotations that passed this review process.

\begin{table}
\centering
\begin{tabular}{lrrrr}
  & \textbf{E} & \textbf{C} & \textbf{A} &\textbf{S} \\ \hline
\#Annotators    & 6377  & 745   & 656   & 31\\
\#Annotations   & 429k  & 426k  & 369k  & 4.8K\\
% \#Unique Words  & 57k   & 78k   & 97k   & 4.8K\\
\#Work Hours    & 10k   & 13k   & 9.0k  & 178\\ \hline
\end{tabular}
\caption{\label{tab:stats} Size of the annotation effort by language. 
}
\end{table}

\section{Dataset Analysis}
\label{sec:analysis}

\subsection{Qualitative}

There are some interesting similarities and differences between language and culture, as discussed in Figure~\ref{fig:teaser}.
% For another example, consider Figure~\ref{fig:familya} where 
There is a considerable inter-annotator agreement (IAA) in the dataset, and there are
also some interesting disagreements.   There is agreement in Figure~\ref{fig:familya} that a mother’s love is universally warm and pleasant. It is an instinct for mothers to be loving, caring and protective of their children.\footnote{\label{footnote:family_caption}
    English caption for Figure~\ref{fig:familya} highlights the cat, whereas
    the Arabic and Chinese focus on the family and do not mention the cat:
    \begin{RLtext} \<أم تجلس مع طفلة صغيرة تنظر إلى طفل يمسك بيدها وتتحدث وتتبادل الحب والمودة.>\end{RLtext} 
    \begin{CJK*}{UTF8}{gbsn}女人看着自己的孩子，让人觉得很开心。\end{CJK*}}
On the other hand, there is a difference in  Figure~\ref{fig:teaser}a.  
All three annotators agree to observe a waterfall though
some mention energy and growth, while others saw horses and wedding veils.

\label{sec:analysis-qualitative}

\subsection{Quantitative}
\label{sec:analysis-quantitative}

Table~\ref{tab:genre} reports multicultural agreement over the 9 emotions\footnote{The 9 emotion classes are: Amusement, Awe,
Contentment, Excitement, Anger, Disgust, Fear, Sadness, and Other} in each genre.

WikiArt classifies artworks into 10 genres,\footnote{The 10 genres are: portrait,
landscape,
genre painting (misc),
religious painting,
abstract painting,
cityscape,
sketch and study,
still life,
nude painting and
illustration.
} as well as 27 styles\footnote{The art styles are:
Abstract Expressionism,
Action painting,
Analytical Cubism,
Art Nouveau Modern,
Baroque,
Color Field Painting,
Contemporary Realism,
Cubism,
Early Renaissance,
Expressionism,
Fauvism,
High Renaissance,
Impressionism,
Mannerism Late Renaissance,
Minimalism,
Naive Art Primitivism,
New Realism,
Northern Renaissance,
Pointillism,
Pop Art,
Post Impressionism,
Realism,
Rococo,
Romanticism,
Symbolism,
Synthetic Cubism and
Ukiyo\_e
}.
Agreement is computed as a log likelihood agreement score, $A = log_2 (Pr(G|D)/Pr(G|U))$,
where $G$ is one of the 10 genres, and $U$ and $D$ are two sets of artworks.  
Let $Pr(G|U)$ be the fraction of artworks in $U$ with genre $G$,
and $Pr(G|D)$ be the fraction of artworks in $D$ with genre $G$.

\begin{figure}[t]
    \centering
    \includegraphics[scale=0.15]{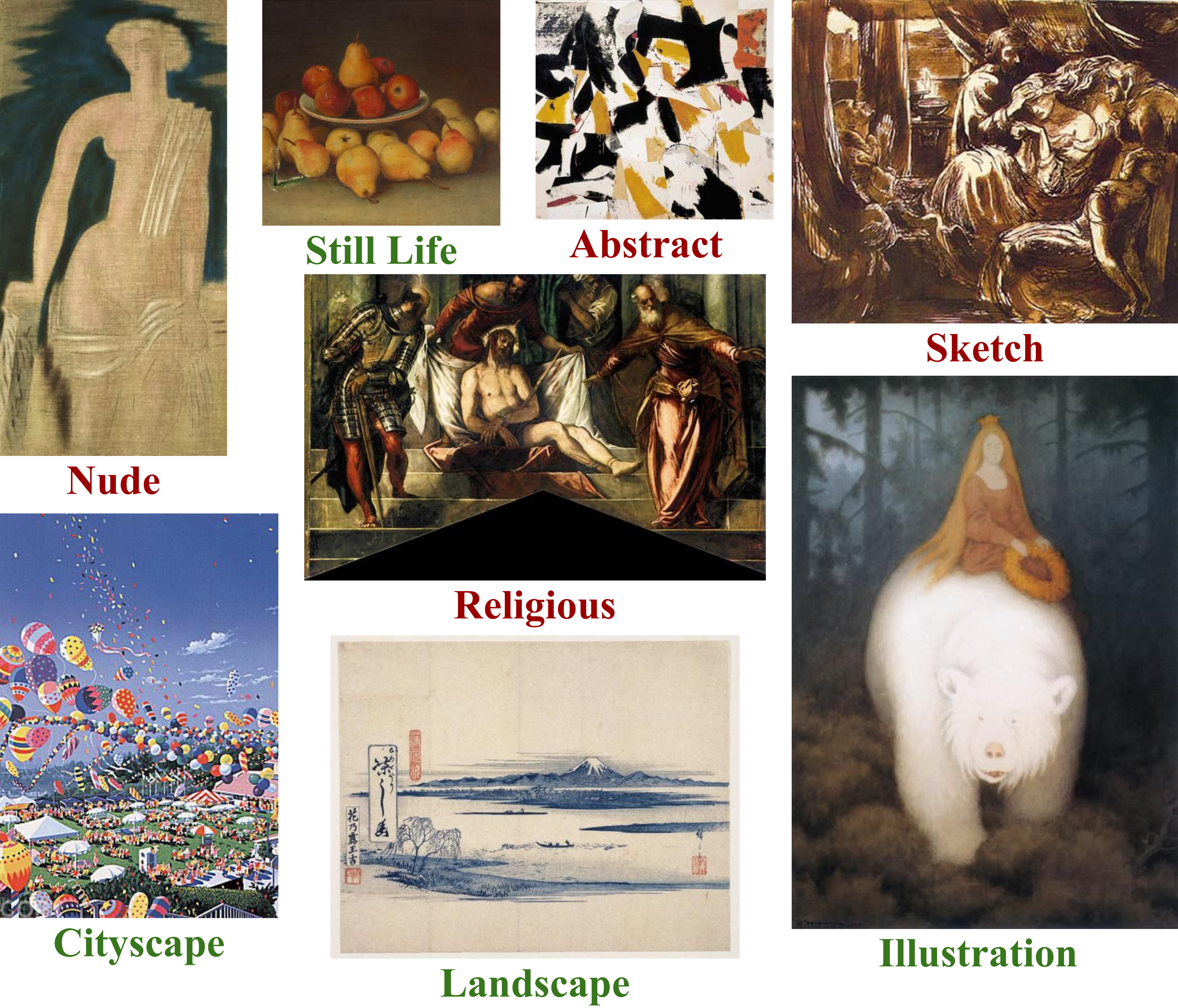}
    \caption{8 artworks with genre. Green indicates high agreement in Table~\ref{tab:genre};  red indicates high disagreement.}
    \label{fig:genres}
    % \vspace{-0.4cm}
\end{figure}

Let $U$ be the universal set
of artworks.  That is, $U$ contains all artworks in \datasetname{} with 5 annotations in each of the 3 languages.
$D$ is a disagreement
set of 2000 artworks.  $D$ was selected by computing Cohen Kappa scores \cite{kappa}\footnote{\url{https://scikit-learn.org/stable/modules/generated/sklearn.metrics.cohen_kappa_score.html}}
for artworks in $U$.  Let $D$ be the 2000 artworks with the most disagreement (based on Kappa).

% \mohamed{what was the threshold?, how big is the set, 2000? some details seems missing in this paragraph}. I believe was addressed 
\begin{table}[t]
\centering
\scalebox{0.9}{
\begin{tabular}{ r | c c | c}
Genre ($G$)& $Pr(G|U)$ &  $Pr(G|D)$ &  $A$ \\ \hline
landscape & 0.206 & 0.097 & -1.08\\
cityscape & 0.071 & 0.036 & -0.98\\
still life & 0.043 & 0.042 & -0.03\\
illustration & 0.029 & 0.029 & -0.01\\
misc & 0.167 & 0.177 & 0.08\\
portrait & 0.217 & 0.233 & 0.10\\
nude & 0.030 & 0.032 & 0.11\\
religious & 0.101 & 0.133 & 0.40\\
abstract & 0.076 & 0.112 & 0.55\\
sketch & 0.061 & 0.109 & 0.85\\
\hline
\end{tabular}
}
\caption{\label{tab:genre} Genre sorted by agreement ($A$).
Most agreement: landscapes; Most disagreement: sketches. %\mohamed{It is not clear how exactly the probabilities in Table 6 are computed from the writing }
}
\end{table}

\begin{figure}[t]
    \centering
    \includegraphics[width=0.38\textwidth]{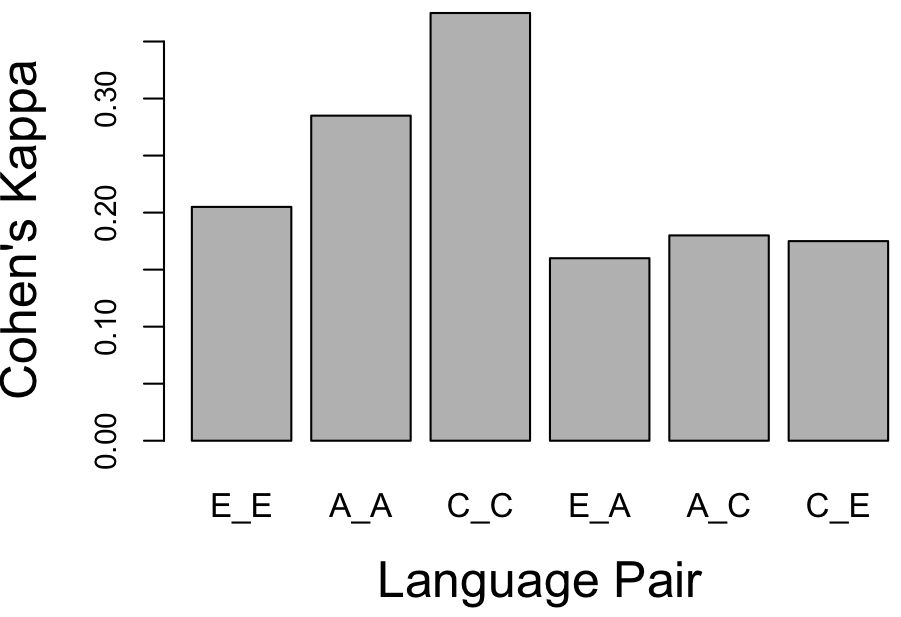}
    \caption{Cohen's Kappa for inter-annotator and cross-annotator agreement. Higher value means more agreement.}
    \label{fig:cohens}
    \vspace{-0.2cm}
\end{figure}

Table~\ref{tab:genre} shows that there is more agreement for some genres (landscapes), and more disagreement
for other genres (sketches).  When the agreement score is near 0, then 
the genre is about equally likely in $U$ and $D$.  This is to be expected for genres near the middle of the list such as misc.  Figure~\ref{fig:genres} shows 8 artworks in genres with high agreement and high disagreement. Figure \ref{fig:cohens} reports the Cohen's Kappa score of annotations from language pairs. Annotators belonging to the same language have higher agreement.

We created $D$ for zero-shot experiments to be reported in
\S\ref{sec:emo-cls}.
The 4.8k Spanish annotations in Table~\ref{tab:stats} are
on the set of $D$ artworks with low IAA (inter-annotator agreement) in ACE (Arabic, Chinese and English).

\section{Emotion Label Prediction}
\label{sec:emo-cls}

Baseline models for two tasks, emotion label prediction and caption generation, will be discussed
in this section and the following section.  These discussions assume
familiarity with deep nets including fine-tuning BERT \cite{devlin2018bert} and
cross language models XLM \cite{conneau2019unsupervised}, as well as HuggingFace \cite{wolf2019huggingface}.

\begin{figure*}[t]
    \centering
    \includegraphics[width=\linewidth]{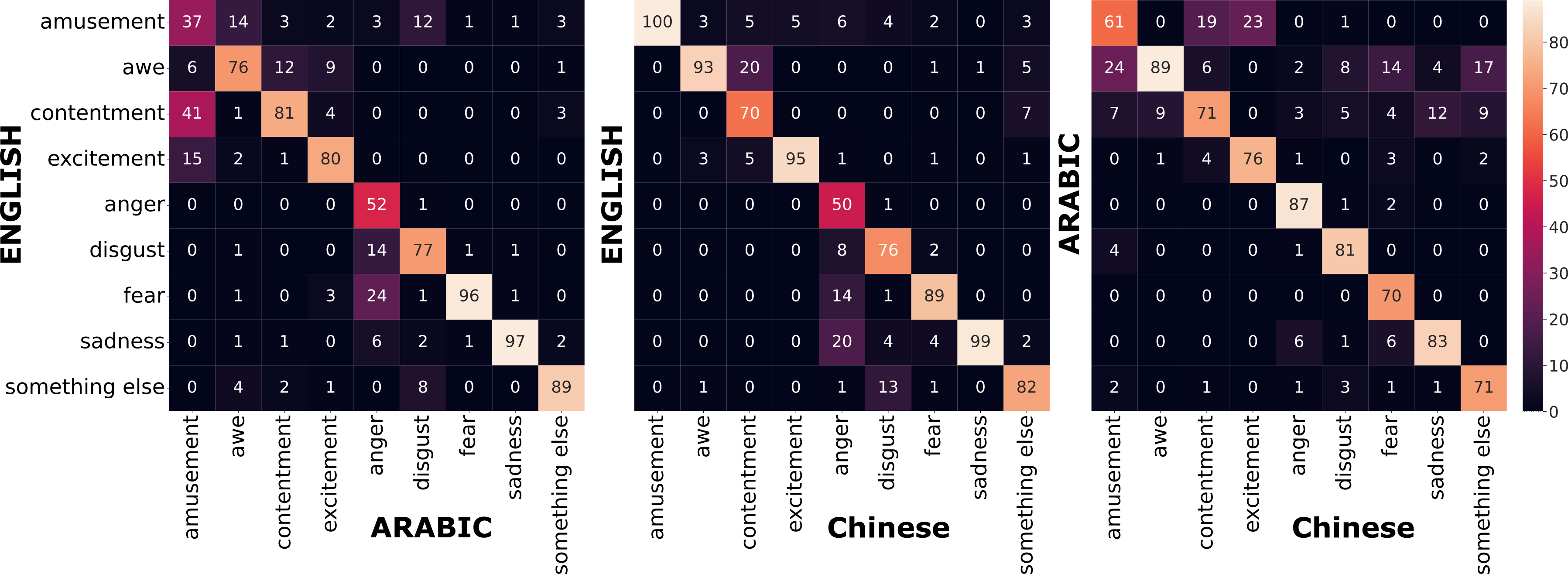}
    \caption{\textbf{Confusion Matrices} The heatmaps show confusion matrices comparing predictions    from the 3-Headed Transformer with ground truth.
    }
    \label{fig:monster_map}
\end{figure*}

\noindent \textbf{Emotion Classification.} Given an input caption, $c$,
we wish to predict an output emotion label, $\hat{e}$, where $\hat{e}$ is one of the 9 emotions.  The model
starts with a pretrained language model, $LM$, and a tokenizer.
The tokenizer converts $c$ into a sequence of $L$ tokens $x$.
The language model converts $x$ into more useful representation, $LM(x) \in R^{L \times d}$, where 
$d$ is the number of hidden dimensions (a property of the LM). Finally, we feed $LM(x)$ into a linear layer to predict the  emotion label, $\hat{e}$.

\noindent \textbf{Majority Baseline.} We use the majority emotion label for each artwork as the predicted emotion for all captions belonging to that artwork. Concretely, each artwork, $I$, has a set of caption-emotion pairs, $S$. The majority classifier outputs the most frequent emotion, $\hat{e}$, in the set $S$ for all of the captions in the set, $c \in S$, 

\noindent \textbf{Language Models.} We finetune 3 models based on BERT (BERT-E, BERT-A and BERT-C),
where BERT-E is tuned for English, and BERT-A is tuned for Arabic and BERT-C is tuned for Chinese. Section \ref{sec:bert} discusses more pretraining and finetuning details.
We also finetune 4 models based on cross language models, XLM-roBERTa \cite{conneau2019unsupervised}, where
XLM-E, XLM-A and XLM-C correspond to English, Arabic, and Chinese  languages, as before.  In addition, we create XLM-ACE by training
on the combination of all 3 languages.

\begin{table}[t]
    \centering
    \scalebox{0.85}{
        \begin{tabular}{l|llll|l}
                        & \multicolumn{5}{c}{Test Set}  \\ \hline
        \small{Backbone}    &   E &   A  &   C &   ACE     &   S \tiny{(0-Shot)}    \\ \hline
        \small{Majority}    &   0.474   &       0.491   &   0.604       &   0.525       &   -       \\ \hline
        \small{BERT-E}     &   0.644   &       -   &   -       &   -       &   -       \\
        \small{BERT-A}     &   -       &   0.558   &   -       &   -       &   -       \\
        \small{BERT-C}     &   -       &       -   &   0.922   &   -       &   -       \\ \hline
        \small{XLM-E}      &   0.662   &   0.345   &   0.781   &   0.606   &   0.513   \\
        \small{XLM-A}      &   0.446   &   0.556   &   0.695   &   0.569   &   0.437   \\
        \small{XLM-C}      &   0.482   &   0.349   &   0.926   &   0.599   &   0.415   \\
        \small{XLM-ACE}     &   \textbf{0.663}   &   \textbf{0.558}   &   \textbf{0.927}   &   \textbf{0.724}   &   0.519   \\
        \hline
        \small{3-Headed-E}    &   0.660   &   0.478   &   0.914   &   0.694   & \textbf{0.529} \\
        \small{3-Headed-A}    &   0.597   &   0.542   &   0.854   &   0.672   & 0.501 \\
        \small{3-Headed-C}    &   0.630   &   0.474   &   0.924   &   0.687   & 0.495 \\
        \small{3-Headed-M}    &   0.653   &   0.498   &   0.917   &   0.700   & 0.525 \\
        \end{tabular}
        }
    
    \caption{\textbf{Emotion Label Classification Baselines.} 
Majority baseline output the most frequent emotion for each artwork.
Models are fine-tuned on BERT and XLM backbones.  Accuracy is best for XLM-ACE.  ``ACE'' combines Arabic (A),  Chinese (C), and English (E). ``M'' stands for mode where the majority vote between the 3 heads is used. For Spanish we evaluate the models without any finetuning (Zero-Shot prediction).}
    \label{tab:emo_cls}
\end{table}

\noindent \textbf{3-Headed Transformer.} Finally, we create a model with XLM-R backbone but replace the single classifier head with 3 classifier heads, one for each of the 3 languages.
% each one composed of a single linear layer and corresponds to one language. 
While training, we feed the captions from each language to the shared backbone and then use the corresponding head to predict an emotion that would ultimately reflect the culture of that language. \citet{geva2021s} analyzed similar multi-headed transformers and showed how the non-target heads can be used to interpret the results of the target head. Similarly, our 3-headed transformer can be used to predict 3 different emotions each one reflecting the culture norms represented in each language. We can then use these predictions to better understand the similarities and differences between cultures. 

\noindent \textbf{Experimental Setup.} We use the base versions of both the BERT and XLM-R models with their default tokenizers from  HuggingFace. We use the standard finetuning procedure where we use the ADAM optimizer to finetune the model for 5 epochs on batches of size 32 with learning rate of $2 \times 10^{-5}$. We use cross entropy as the loss function for updating the full model parameters, including the transformer backbone. We follow the standard ArtEmis \cite{achlioptas2021artemis} splits introduced in \cite{youssef2022artemis2} and adopt them for both Arabic and Chinese datasets.  The same training and testing images are used in all cases. For BERT models, we only evaluate on the same language as the training set because BERT tokenizers are language specific.

\noindent \textbf{Baseline Results.} Table~\ref{tab:emo_cls} reports accuracy for several BERT/XLM models.
There are 4 test sets,  one for each language, plus ACE (a combination of 3 languages).
XLM models perform better than BERT,
because there is no data like more data, as well as the cross language setup used during pretraining.
Interestingly, scores on the Chinese test set are higher than for English and Arabic, suggesting that Chinese captions are easier to classify.
Finally, notice that XLM-ACE (XLM trained on 3 languages) outperforms other conditions,
showcasing benefits of multiple languages.
Note that XLM-ACE even outperforms matching conditions, where training language = test language.

\noindent \textbf{3-Headed Transformer Analysis.} Although the 3-Headed transformer did not improve accuracy, the 3 classification heads are useful for error analysis.  We feed the entire \datasetname{} dataset to the model and predict 3 $\hat{e}$ values, one for each head/language.  Confusion matrices are reported in Figure~\ref{fig:monster_map}. There is more agreement on negative emotions,
and less agreement on  positive emotions.

We are interested in large off-diagonal values in Figure~\ref{fig:monster_map},
especially between positive and negative emotions. 
For example, Arabic \textbf{disgust} is often confused with English \textbf{amusement}.

Upon further investigation, we found nude paintings contributed ${\sim}15\%$ of these confusions.
Explicit content and alcohol are frowned upon in some Arabic speaking communities, as
illustrated by the second and third rows of Table~\ref{tab:emo_cls_samples}, where the label is positive in English and Chinese,
but not in Arabic.

Religious symbols are also associated with large off-diagonal values in confusion matrices.  
The first row in Table~\ref{tab:emo_cls_samples} mentions Jesus and how a beautiful girl holds his cross and stomps on the devil. 
The annotation is positive (\textbf{awe}) in English and Arabic, but negative (\textbf{fear}) in Chinese.
In China, the cross holds less meaning, and stomping on the devil is more scary than reassuring.
Many symbols are associated with religion, holidays and legends that mean more in some places
than others.\footnote{Dragons are positive in East, but negative in West.}

While there are a few off-diagonal cells with large values, most of the large values in the confusion matrices
are on the main diagonal.  That is, the similarities across languages tend to dominate the differences.
Consider the last row in Table~\ref{tab:emo_cls_samples}, which 
receives a positive label (\textbf{contentment}) in all 3 languages.  Babies make people feel happy (nearly) everywhere.
In this case, all 3 heads of our 3-headed transformer predict positive labels for this caption.
For training models across multiple languages, similarities across languages may be more
useful than differences.

\begin{table}[!t]
    \centering
    \scalebox{0.8}{
        \begin{tabular}{p{0.65\linewidth}|ccc}
            % \hline
                        &   \multicolumn{3}{c}{Transformer Head} \\
            % \hline
            Input Caption (Gloss)     &   E      &   A      &   C  \\ 
            \hline  
            [A] A beautiful girl holding a Jesus cross stomping on the devil   &   Awe &   Awe &   Fear \\
            \hline  
            [E] The woman on the ground isn't wearing any clothes   &   Amu. &   Dis. &   Amu. \\
            \hline  
            [E] The man looks like he's drunk since his expression is so wired out   &   Amu. &   Sad  &   Exc. \\
            \hline  
            [C] Countless babies have descended into the world, giving life to the world and making people feel happy.   &   Cont. &   Cont. &   Cont. 
        \end{tabular}
    }
    \caption{  \label{tab:emo_cls_samples} \textbf{Predictions from 3-Headed Transformer}: The input is a caption in Arabic (A), Chinese (C) or English (E).  The first column shows the language and a gloss.  The last three columns
    show predictions for each head (with interesting differences across heads).}

\end{table} 

\paragraph{Zero-Shot Evaluation.} We use Spanish annotations in \datasetname{} to evaluate
models mentioned above in a zero-shot setting.  The last column in Table~\ref{tab:emo_cls} reveals two interesting relations:
\begin{enumerate}
 \setlength{\itemsep}{0pt}
    \setlength{\parskip}{0pt}
    \setlength{\parsep}{0pt}
    \item 3-Headed-E $>$ XLM-ACE
    \item 3-Headed-E $>$ 3-Headed-A $>$ 3-Headed-C 
\end{enumerate}
The first relation suggests that 3-Heads may not perform as well as XLM when there is plenty of data,
but 3-Heads may have advantages in low-resource and zero-shot settings.  3-Heads
are better for capturing interactions between languages.

The second relation 
suggests that language transfer may be more effective across some language pairs than others.
Historically, Spanish and English are both relatively close Indo-European languages,\footnote{\url{http://www.sssscomic.com/comicpages/196.jpg}}
compared to Semitic languages such as Arabic.
There has been much less contact \cite{thomason2001language}
between those languages and Chinese.

\section{Affective Caption Generation}
\label{sec:neural-speakers}

The previous section described baseline models for the first task: label prediction.
This section will describe baseline models for the second task: affective caption generation.

To this end, we follow \citet{achlioptas2021artemis} and train two affective captioning models: Show, Attend, and Tell (SAT)~\cite{xu2015show} and Meshed Memory Transformer ($M^2$)~\cite{cornia2020meshed}. We use \emph{Affective Captioning Models} to refer to captioning models that generate affective captions.  These captions connect the dots between input paintings and emotions.

SAT is a LSTM \cite{hochreiter1997long} based captioning model with an attention module, it consists of a visual encoder and a text decoder. The visual encoder extracts visual features from an input image. The decoder then uses a stack of an attention module and LSTM recurrent unit to generate a caption autoregressively. 
$M^2$ is a transformer based model \cite{vaswani2017attention} which utilizes a pretrained Faster-RCNN \cite{ren2015faster} object detector to extract visual region features. These features are used as an input sequence to a multi-layer attention based encoder. $M^2$ differs from basic transformers by feeding the encoded features from all encoder layers to the cross attention module in each decoder's layer. In order to include Emotion and Language grounding, we use a simple embedding layer to convert the emotion and language labels into feature vectors and then concatenate them to the visual features. 

\paragraph{Experimental Setup.} For both models, we use the default parameters proposed in \cite{achlioptas2021artemis}. We train four different versions of each model, three versions are trained on English, Arabic, and Chinese only datasets, while the fourth version is trained on the three languages combined. We then test all the models on all the languages. In order to allow the models to work on an arbitrary languages during testing, we create our custom tokenizer which is based on xlm-roberta-base tokenizer from HuggingFace.  The available tokenizer has a vocabulary of size 200K tokens which makes the training inefficient. To mitigate this, we use the same \emph{xlm-roberta-base}\footnote{https://huggingface.co/xlm-roberta-base} tokenizer training strategy to create a tokenizer with 60K vocabulary size on \datasetname{}.% \mohamed{why it became 60k instead of 200k? because the training with 200K is inefficient.}. 

\begin{table}[t]
    \centering
    \scalebox{0.71}{
        \begin{tabular}{ll|llll|llll}
            \multicolumn{2}{c}{} & \multicolumn{4}{|c|}{SAT} & \multicolumn{4}{c}{$M^2$} \\ 
            \multicolumn{2}{c|}{Test Set} & E & A & C & ACE & E & A & C & ACE \\ \hline
            \multirow{4}{*}{E} &$B_4$ & 6.2 & 0 & 0 & 6.9  & \textbf{8.7} & 0 & 0 & 8.1     \\ 
                                & $M$ & 13.9 & 0 & 0 & \textbf{14.2}  & 12.9 & 0 & 0 & 12.4     \\ 
                                & $R$ & 26.5 & 0 & 0 & 26.5  & \textbf{28.0} & 0 & 0 & 27.4     \\ 
                                & $C$ & 6.4 & 0 & 0 & 6.3  & 9.2 & 0 & 0 & \textbf{9.4}     \\  \hline
            \multirow{4}{*}{A} &$B_4$ & 0 & 3.1 & 0 & 3.2  & 0 & 3.5 & 0 & \textbf{3.7}     \\ 
                                & $M$ & 0 & 30.2 & 0 & 30  & 0 & \textbf{30.9} & 0 & 30.7     \\ 
                                & $R$ & 0 & 15.4 & 0 & 15.4  & 0 & 15.1 & 0 & \textbf{15.5}     \\ 
                                & $C$ & 0 & 7.7 & 0 & 7.7  & 0 & 7.8 & 0 & \textbf{8.0}     \\  \hline
            \multirow{4}{*}{C} &$B_4$ & 0 & 0 & \textbf{11.9}  & 10.9  & 0 & 0 & 8.3 & 8.7     \\ 
                                & $M$ & 0 & 0 & \textbf{16.1}  & 15.8  & 0 & 0 & 15.1 & 14.6     \\ 
                                & $R$ & 0 & 0 & \textbf{34.3}  & 33.6  & 0 & 0 & 31.1 & 31.1     \\ 
                                & $C$ & 0 & 0 & \textbf{9.5}  & 8.5  & 0 & 0 & 8.9 & 7.8     \\  \hline
        %   \multirow{4}{*}{ACE}&$B_4$ & 06.5 & 02.4  &   &   & 08.5 & 02.2 & 0 & 08.7     \\ 
\multirow{4}{*}{\rotatebox[origin=c]{90}{ACE}}&$B_4$ & 6.0 & 0  & 11.3  & 9.6 & 8.9 & 3.9 & 8.3 & \textbf{27.4}     \\ 
                                & $M$ & 13.5 & 0.42  & 15.2  & 10.5  & 11.8 & \textbf{30.8} & 14.8 & 21.2     \\ 
                                & $R$ & 28.9 & \textbf{94.8}  & 33.3  & 51.8  & 27.6 & 45.1 & 30.5 & 32.1     \\ 
                                & $C$ & 2.4 & 0.06  & 3.0  & 2.0  & 3.1 & \textbf{14.6} & 3.1 & 5.6     \\  \hline
        \end{tabular}
    }
    % \vspace{2mm}
    \caption{\textbf{Affective Captioning Baseline.} SAT and $M^2$ are trained on English (E), Arabic (A), Chinese (C), and all languages (ACE). The trained models are evaluated on a test set from each language as well as a combined test set. For metrics, we use BLEU-4 ($B_4$), METEOR (M), ROUGE (R), and CIDEr ($C$). Each row corresponds to a test set in a particular language. Meanwhile, columns correspond to model trained on a given language.}
    \label{tab:caption_baseline}
    \vspace{2mm}
\end{table}

\paragraph{Results.} We report the results of our baseline models in Table~\ref{tab:caption_baseline}. Models trained using all the languages perform very similarly to their language specific counterparts on every metric except for the Chinese language. This provides additional evidence that English and Arabic speaking cultures are more closely related to one another than either is to Chinese ones. In other words, English captioning models do not lose much performance when Arabic data is added to the training set and vice versa. On the other hand, Chinese models suffer when such data is added. Moreover, we also observe that for models trained on single languages, the scores on the combined test set is proportional to the language specific test sets. 

\section{Conclusion}
\label{sec:conclusion}

This paper introduced \datasetname{},
a multilingual dataset and benchmark on WikiArt images with more than 1.2M captions and emotion labels. The benchmark has diverse emotional experiences constructed over different cultures, and  communicated in four languages (English, Chinese, Arabic, and Spanish).  We found more agreement for some genres
such as landscapes and more disagreement for other genres such as sketches.
These differences are interesting and important, and far from random.
Annotations for trees in Figure~\ref{fig:teaser}c are labeled as sadness in English and Chinese
but contentment in Arabic.  People are likely to feel more comfortable
with what they know.  People raised in countries with lush forests are likely to prefer
that, whereas people brought up in less humid environments are likely to prefer that.

Towards building more socially and multi-culturally aware AI,  we created baseline models for two tasks on \datasetname{}: (1) emotion label prediction and (2) affective caption generation. For emotion label prediction, our best baseline model trained XLM on a combination of training data from all three languages (XLM-ACE).  We also created 3-headed transformers,  training three heads for three languages (Arabic, Chinese, and English) at the same time.
The performance of this model is close to XML-ACE, but generalizes better in a zero-shot
experiment on Spanish. For the caption generation task, we trained two models on \datasetname{}, SAT and $M^2$.  For English and Arabic,
models on all three languages have a similar performance to language specific models, but for Chinese, it is best to train
without the other languages since the performance drop is significant. 

We hope our benchmark and baselines will help ease future research in visually-grounded language models that can communicate affectively with us. In addition, \datasetname{} can provide empirical examples of cross-cultural similarities and differences. Sociologists and Cultural Psychologists may formulate hypotheses and conduct field studies based on \datasetname{}. 
Data, code, and models are publicly available at \url{www.artelingo.org/}.

\section{Limitations}
\label{sec:limitations}
\datasetname{}'s artworks are extracted from WikiArt. Although \datasetname{} is diverse in language and culture, it inherits WikiArt's bias toward western artworks as discussed in Table~\ref{tab:wikiArt_diversity} in \S\ref{sec:representation}.  There is room to improve
the representation of certain regions of the world. 
Due to globalisation, people tend to follow similar trends around the world, causing others to follow their lead (for better and for worse).

Many cultures, such as Arabic, do not have a rich heritage of oil paintings. Instead, they have other forms of Art like poetry and calligraphy. Such art forms are interesting to study on their own, but mixing them with paintings is not obvious. Based on the original ArtEmis dataset, we chose WikiArt with the intent to be a continuation of their work. Also, artworks are more accessible and can be interpreted easier by different cultures compared to poetry and other art forms.

The addition of affective captions for Arabic,  Chinese, as well as a small set of Spanish is a step
toward cultural diversity. However, more than four regions and  languages are indeed needed to cover the world. Scalability can be a challenge. However, we hope that progress can be accelerating by developing affective vision and language models that can learn with limited data for each additional language by distilling knowledge from language-only models as in \cite{chen2022visualgpt,alayrac2022flamingo}. 

\datasetname{} was also collected through AMT's online platform\footnote{Chinese data collection was done on AMT as a tool with help from Baidu who helped recruit the human participants from China}. This suggests that the workers are familiar with technology and social media, imposing an influence on the data.
Social media influences many concepts such as: trending news, and standards, which may lead to the presence of similarities between cultures. There have been, of course, other concerns about the use of AMT and the so-called ``gig'' economy and workers' rights.

\section{Acknowledgements}
The authors would like to thank Baidu for soliciting Chinese annotators, all the annotators for their effort in the data collection, Eric Macedo Esparza for reviewing the Spanish dataset, and the anonymous reviewers for their valuable comments. We also would like to thanks all the middle eastern universities, mainly in Egypt,  who contributed to collecting the Arabic version. 

This work was supported by King Abdullah University of Science and Technology (KAUST), under Award No. BAS/1/1685-01-01. 

\bibliography{anthology,custom}
\bibliographystyle{acl_natbib}

\clearpage
% \onecolumn
\section{Appendix}
\label{sec:appendix}

\subsection{GitHub Repo}
\label{sec:github}
You can find the dataset and more visuals in \url{artelingo.org} or our github repo \url{github.com/Vision-CAIR/artelingo}

\subsection{Pretrained BERT models}
\label{sec:bert}
In the emotion prediction experiment, we finetune pretrained BERT models. For each language, we use a BERT model pretrained only on that language. In particular, we use ``bert-base-uncased''\footnote{\url{https://huggingface.co/bert-base-uncased}} for English; ``CAMeL-Lab/bert-base-arabic-camelbert-mix''\footnote{\url{https://huggingface.co/CAMeL-Lab/bert-base-arabic-camelbert-mix-ner}} for Arabic; and ``bert-base-chinese''\footnote{\url{https://huggingface.co/bert-base-chinese}} for Chinese.

Language specific models are finetuned on subsets of ArtELingo having captions written in the same language. On the other hand, multilingual models are pretrained ``XLMroBERTa''\footnote{\url{https://huggingface.co/xlm-roberta-base}} and they are finetuned on the whole of ArtELingo.

For each model, we finetune the pretrained model for 5 epochs. We use an ADAMW optimizer \footnote{\url{https://huggingface.co/docs/transformers/main_classes/optimizer_schedules##transformers.AdamW}} with a learning rate of $2\times 10^{-5}$ with a linear schedule\footnote{\url{https://huggingface.co/docs/transformers/main_classes/optimizer_schedules##transformers.get_linear_schedule_with_warmup}}. We use cross-entropy as the loss function\footnote{\url{https://pytorch.org/docs/stable/generated/torch.nn.CrossEntropyLoss.html}}. Please check our GitHub repo for all of the implementation details\footnote{\url{github.com/Vision-CAIR/artelingo}}.

\subsection{Ethical Concerns}
\label{sec:ethics}
We received approval for the data collection from KAUST Institutional Review Board (IRB). The IRB requires informed consent; in addition, there are terms of service in AMT. We respected fair treatment concerns from EMNLP (compensation) and IRB (privacy). We compensated the workers well above the minimum wage ($<$\$1 USD/hour in Egypt and \$2.48 USD/hour in China). We paid our workers \$0.07 USD per completed task. Each task takes on average 50 seconds to complete. In addition, we paid bonuses (mostly 30\%) to workers who submitted high-quality work.

The workers were given full-text instructions on how to complete tasks, including examples of approved and rejected annotations (please refer to \S\ref{user_interface}). Participants’ approvals were obtained ahead of participation. Due to privacy concerns from IRB, comprehensive demographic information could not be obtained. 

\clearpage
\onecolumn
\subsection{User Interfaces}
\label{user_interface}
% \begin{multicols}{1}
\begin{figure*}[h!]
    % \hspace{-15cm}
    \centering
        \includegraphics[width=\textwidth]{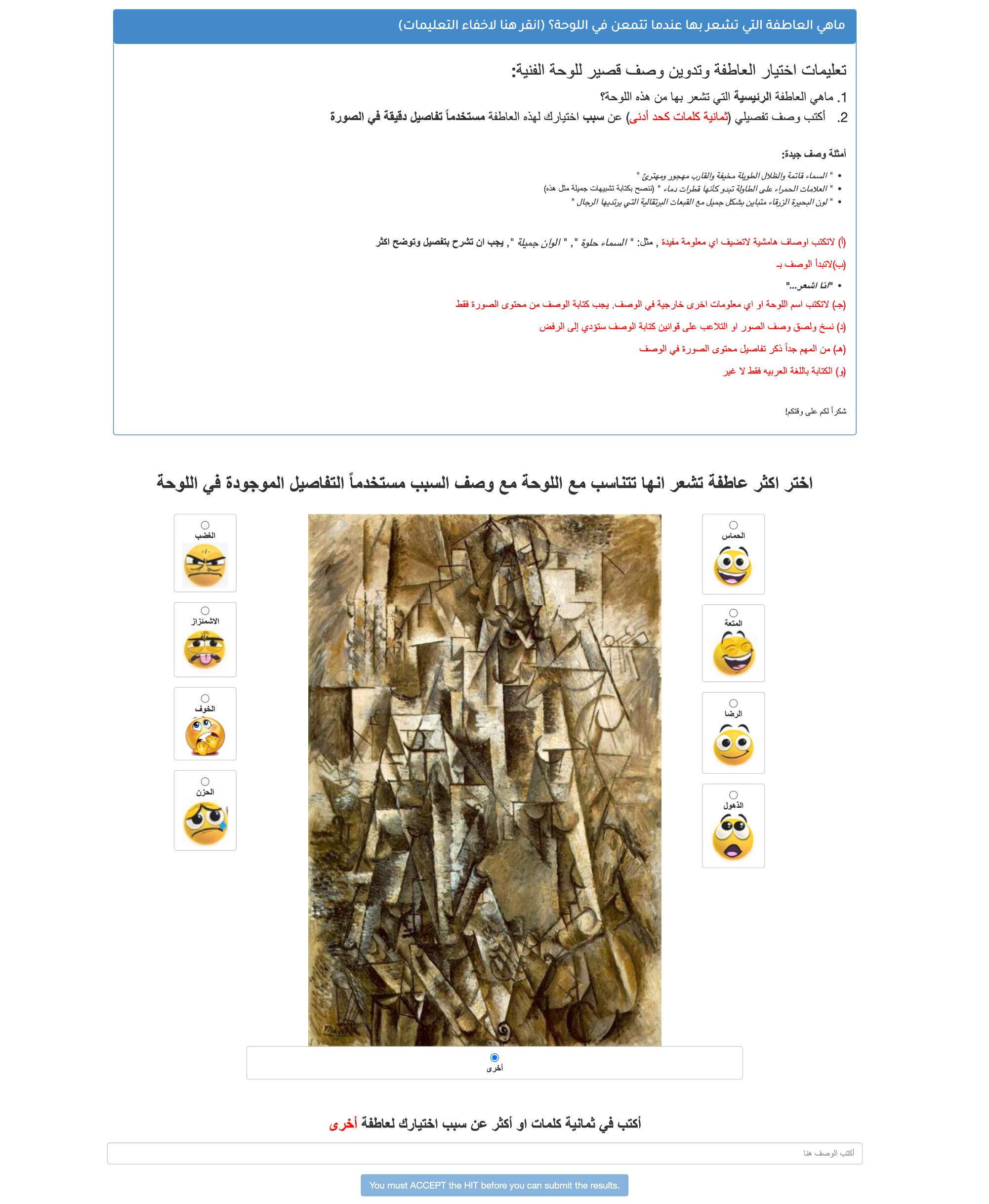}
         \caption{Arabic Interface}
         \label{fig:ar_interface}
\end{figure*}
% \end{multicols}

\begin{figure*}[t]
    \centering
         \includegraphics[width=\textwidth]{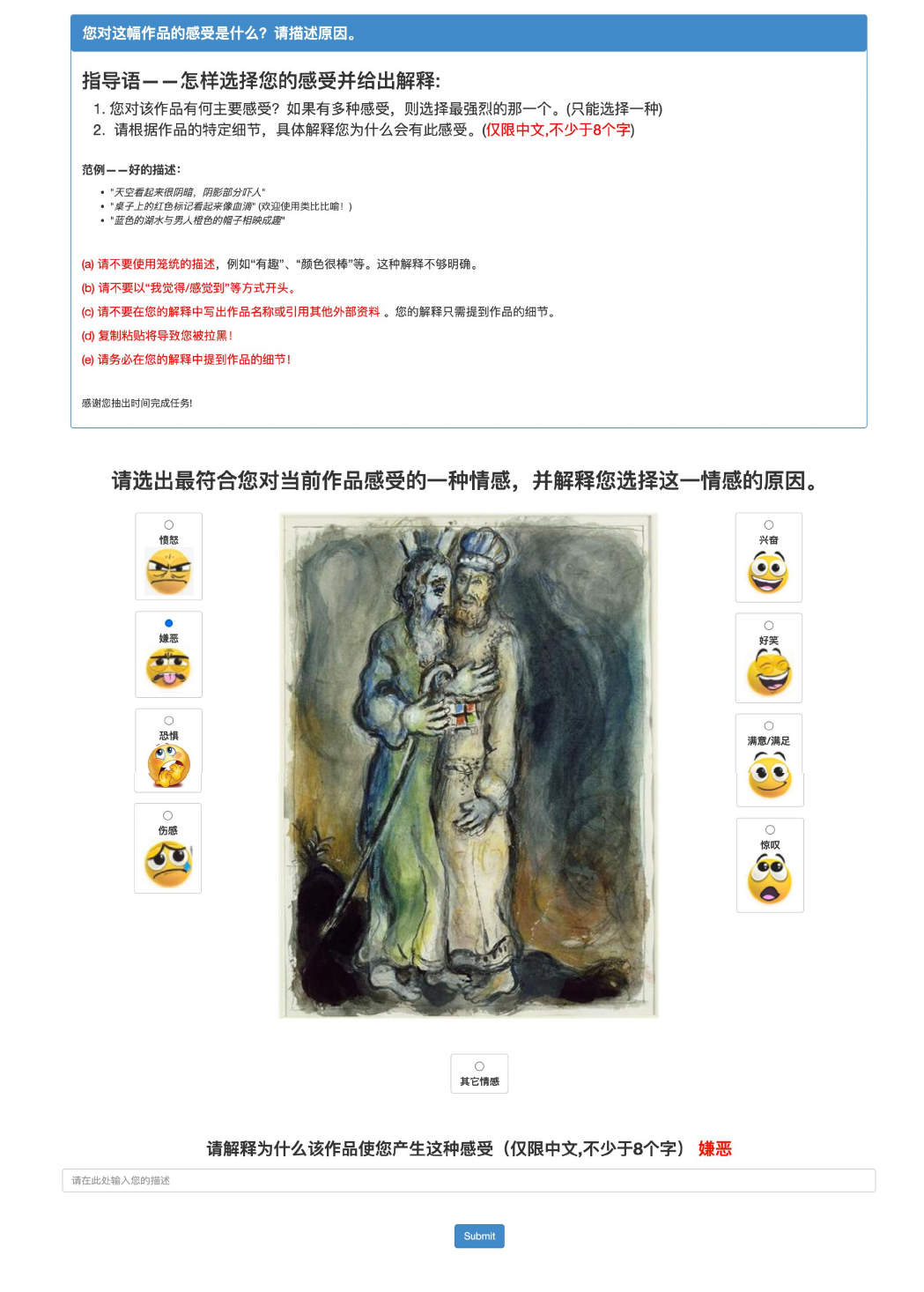}
         \caption{Chinese Interface}
         \label{fig:ch_interface}
\end{figure*}

\begin{figure*}[t]
    \centering
         \includegraphics[width=\textwidth]{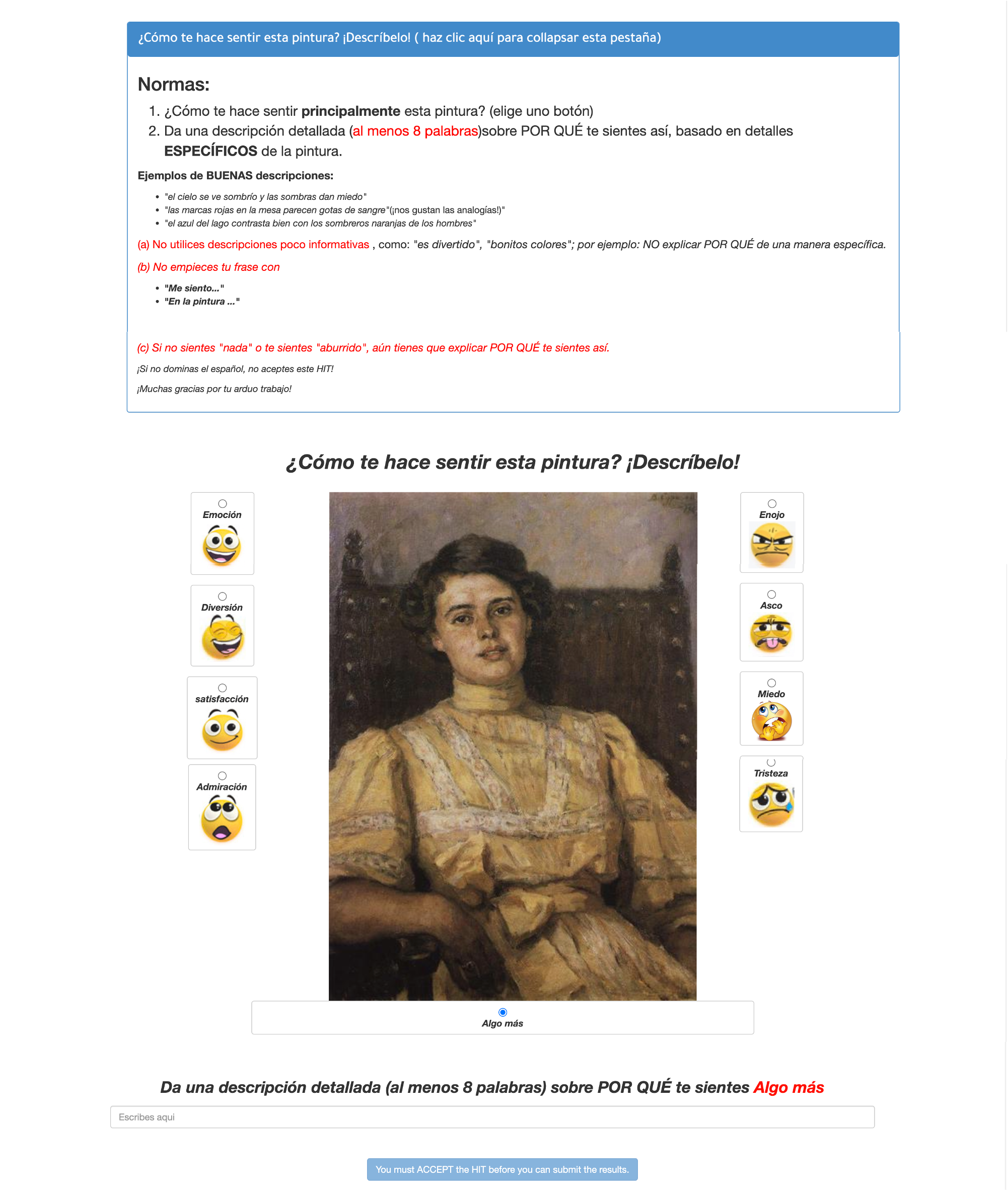}
         \caption{Spanish Interface}
         \label{fig:sp_interface}
\end{figure*}

\end{document}